\documentclass{article}
\usepackage{iclr2027_conference,times}
\usepackage{iftex}
\ifPDFTeX
  \usepackage[T1]{fontenc}
\else
  \usepackage{fontspec}
\fi

\AtBeginDocument{}
\usepackage{amsmath,amssymb}
\usepackage{enumitem}  %
\usepackage{booktabs} %
\usepackage{etoolbox}
\AtBeginEnvironment{table}{\setlength{\belowcaptionskip}{4pt}}
\usepackage{colortbl} %
\usepackage{multirow, multicol} %
\usepackage{pifont} %
\usepackage{xcolor}

\newcommand{\cmark}{\ding{51}}
\newcommand{\xmark}{\ding{55}}

\definecolor{tabfirst}{rgb}{1,0.7,0.7}
\definecolor{tabsecond}{rgb}{1,0.85,0.7}
\definecolor{tabthird}{rgb}{1,1,0.7}
\definecolor{darkred}{rgb}{0.6,0.16,0.25}
\definecolor{heavygray}{gray}{0.90}

\usepackage{tikz}
\usetikzlibrary{positioning,arrows.meta,calc,fit,backgrounds,decorations.pathreplacing}
\definecolor{apviolet}{HTML}{6A51B9}
\definecolor{apslate}{HTML}{7F8CA3}
\definecolor{apblue}{HTML}{A1B2BA}
\definecolor{appink}{HTML}{D3A0B3}
\definecolor{apgreen}{HTML}{458863}
\definecolor{aptaupe}{HTML}{B9ADA1}
\definecolor{apink}{HTML}{071A4A}

\usepackage{graphicx,array,placeins,float}
\usepackage{flafter}  %
\usepackage[colorlinks=true,linkcolor=black,citecolor=blue,urlcolor=blue]{hyperref}
\usepackage{url}
\newcommand{\best}[1]{\textbf{#1}}
\newcommand{\model}{PatchWAM}
\title{An Action Is Worth One Patch: Unified\\World--Action Modeling with \model{}}
\author{Tianheng Wang$^{1,\dagger}$ \quad Zhou Xie$^{2}$ \quad Heng Jia$^{3}$\\
\bfseries Jianhua Xu$^{4}$ \quad Tong Zhang$^{5}$ \quad Kaicheng Yu$^{1,4,\ast}$\\[2pt]
\mdseries $^{1}$Westlake University \quad $^{2}$Lanzhou University \quad $^{3}$Zhejiang University \quad $^{4}$Awomo\\
\mdseries $^{5}$University of Chinese Academy of Sciences\\[2pt]
\mdseries\small\texttt{wangtianheng@westlake.edu.cn, xzhou2024@lzu.edu.cn, 12221194@zju.edu.cn}\\
\mdseries\small\texttt{xujianhua@westlakedi.com, tongzhang@ucas.ac.cn, kyu@westlakedi.com}\\[2pt]
\mdseries\small $^{\ast}$Corresponding author \quad $^{\dagger}$Work done during an internship at Awomo
}

\iclrfinalcopy
\begin{document}
\maketitle
\lhead{}  %
\begin{abstract}
Generative visual models offer a foundation for learning representations of physical dynamics, yet their extension to continuous control raises a fundamental question: do visual prediction and action generation require separate computational pathways?
Existing approaches usually introduce trainable action heads or separate action experts to bridge low-dimensional states and high-dimensional visual representations.
In this work, we explore whether the visual backbone’s existing capacity can also support control when actions are expressed in a compatible representation.
Thus, we introduce \model{} (Patch World-Action Model), which treats continuous actions as another type of patch through a fixed mapping called Action-as-Patch.
This allows a single model to predict both how the robot should move and what the scene may look like afterward.
Visual prediction and action generation become parts of the same generative process, without a dedicated action head or separate action expert.
Experiments with subsampled training windows show gains over a matched dual-expert control, while benchmark evaluations reach 91.8\% success rate on LIBERO-Plus and 96.12\% on RoboTwin 2.0 in a full-data setting with additional augmented demonstrations.
More broadly, the result suggests that capability need not be added where it can be inherited: the constraint on extending a generative backbone is the interface a new signal is written in, not the capacity to model it.
\end{abstract}

\begin{figure}[t]
    \centering
    \includegraphics[width=\textwidth]{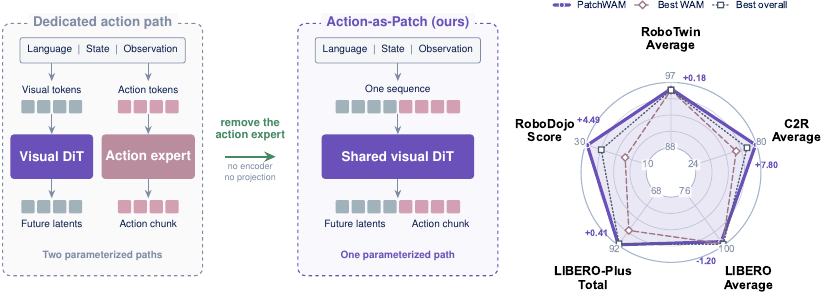}
    \caption{
        \textbf{Does action prediction need a dedicated DiT?}
        \textbf{Left:} \model{} replaces the separate action expert with fixed Action-as-Patch encoding and a shared visual DiT, so the only difference between the two panels is the removed action expert; a filled block is a parameterized network and nothing else is filled.
        Figure~\ref{fig:architecture} gives the codec and the joint denoising in full.
        The left half abstracts the dual-expert control used in our comparisons rather than describing every dual-expert VLA.
        \textbf{Right:} success rates on five benchmarks, with the Score for RoboDojo; each spoke spans its own leaderboard range because absolute values are not comparable across benchmarks, and the number beside each spoke gives the difference between \model{} and the strongest listed method.
        Figure~\ref{fig:benchmark-summary} gives the success rate of \model{} relative to the WAM SoTA on each benchmark, and Tables~\ref{tab:robotwin-full}, \ref{tab:robotwin-c2r}, \ref{tab:libero-results}, \ref{tab:libero-plus-results}, and~\ref{tab:robodojo-comparison} give the full comparisons.
    }
    \label{fig:intro-overview}
\end{figure}
\section{Introduction}
\label{sec:intro}
Robotic actions and their visual consequences describe complementary aspects of the same interaction: actions specify the robot's motion, while future images capture the resulting scene configuration \citep{ha2018worldmodels,kim2026cosmospolicy,alzayer2026maskedvisualactions}.
Pretrained image-generation and editing models provide visual priors over objects and spatial relationships, together with a trained mechanism for conditional generation of continuous latent tokens \citep{rombach2022ldm,esser2024sd3,bflklein,zhang2026imagewam}.
Action prediction can be formulated using the same denoising objective, conditioned on the same observations and task instruction \citep{black2024pi0,zhu2025uwm}.
This computational overlap motivates us to investigate whether continuous control requires a separate action expert \citep{black2024pi0,intelligence2025pi05,motubrain2026,cai2026ahawam,gigaworld2026policy05,chen2026abotm05,wang2026openwam}, or whether the existing visual generative pathway can be adapted to predict actions directly.
A fixed action representation allows us to test this possibility without introducing a learned action-specific interface.

We introduce \model{}, a Patch World-Action Model (WAM; \citealp{ye2026dreamzero,zhang2026imagewam,shen2026wamsurvey,wang2026wamfrontier}) built around \emph{Action-as-Patch} (AP), a fixed codec that maps each action step to a token in the visual latent space (Figure~\ref{fig:intro-overview}, left).
For the action spaces considered here, the action dimension is smaller than the visual token width, allowing an information-preserving mapping through repetition and zero-padding.
A fixed decoder recovers the encoded action by group averaging.
Action tokens therefore use the backbone's existing visual processing path, including its learned input and output projections, without a trainable action encoder, action-specific output projection \citep{nvidia2025gr00tn1,pai2025mimicvideo,wang2026qwenvla}, or separate action expert.
Throughout this paper, a \emph{dedicated action head} denotes a trainable action-specific prediction module beyond deterministic decoding.

Conditioned on language, proprioception, and current images, \model{} jointly predicts an action chunk and a future visual endpoint.
During training, action and future-image tokens are perturbed with independent Gaussian noise at a shared noise level and supervised by a joint flow-matching objective \citep{lipman2022flow}.
A shared diffusion transformer (DiT; \citealp{peebles2023dit}) denoises both token groups, allowing interaction between visual and action predictions.
At inference, the fixed decoder converts the resulting action tokens into commands for receding-horizon execution; decoding the future visual latent into RGB is unnecessary.

We evaluate this construction against a dual-expert control under matched data and optimization settings on RoboTwin 2.0 \citep{chen2025robotwin2}.
With one retained training-window start out of every twenty, \model{} achieves 88.0\% average success compared with 78.4\% for the control, a gain of 9.6 percentage points with fewer trainable parameters (Section~\ref{sec:sparse}).
The result establishes the viability of the fixed interface in this sampling regime, although the comparison does not isolate the effects of parameter sharing and attention organization.
Additional evaluations characterize performance under different training and generalization protocols.
\model{} reaches 96.12\% average success on RoboTwin 2.0 with clean, randomized, and additional augmented demonstrations, and 91.8\% on LIBERO-Plus \citep{fei2026liberoplus} with augmented training (Figure~\ref{fig:intro-overview}, right).
In clean-to-random transfer \citep{robotwinleaderboard2026}, it achieves 66.72\% randomized-scene success and a 79.14\% clean/random mean; the randomized score remains below the strongest listed baseline.
Original LIBERO \citep{liu2023libero} and RoboDojo \citep{chen2026robodojo} evaluations further characterize the model's capabilities and limitations (Section~\ref{sec:benchmarks}).
These benchmark comparisons involve different data regimes and are interpreted separately from the matched architecture control.
Because image and action latents are updated at every denoising step, the reduction in trainable parameters does not imply lower inference latency (Section~\ref{sec:cost}); several WAMs reduce this cost by skipping future prediction at test time or by distilling the solver \citep{yuan2026fastwam,ye2026gigaworld,akbari2026flashwam}.

Our contributions are threefold:
\begin{itemize}[leftmargin=1.5em,itemsep=2pt,topsep=4pt]
\item \textbf{A fixed action interface.} An information-preserving codec represents continuous actions as visual-space tokens, enabling action prediction without a dedicated trainable action head.
\item \textbf{Joint visual and action prediction.} A shared transformer predicts future-image and action tokens through a joint flow-matching objective with explicit conditioning and attention structure.
\item \textbf{Empirical evaluation.} A matched dual-expert comparison demonstrates effective control with the fixed interface, complemented by five evaluation settings covering manipulation and perturbation robustness under distinct training protocols.
\end{itemize}

\section{Related Work}
\label{sec:related-work}
\paragraph{Vision--language--action models.}
Vision--language--action (VLA) models build robot policies on top of pretrained vision--language models.
Certain VLAs, such as OpenVLA, represent actions as discrete tokens \citep{kim2024openvla}.
Most recent VLAs instead generate continuous actions through an additional module, either a learned action head \citep{nvidia2025gr00tn1,wang2026qwenvla} or a separate action expert \citep{black2024pi0,intelligence2025pi05}.
These models inherit semantic knowledge from vision--language pretraining, but their action-generation modules typically learn task-specific physical dynamics primarily from robot data \citep{pai2025mimicvideo}.
This limitation has motivated policies grounded in generative visual models.

\paragraph{World action models.}
World action models (WAMs) derive robot policies from video or image generation models and predict future observations jointly with actions \citep{wang2026wamfrontier,shen2026wamsurvey,ye2026dreamzero,zhu2025uwm}.
Numerous WAMs nonetheless retain a separate pathway for actions, in the form of a dedicated action stream or expert \citep{motubrain2026,cai2026ahawam,gigaworld2026policy05,chen2026abotm05}.
A complementary line of work reduces the cost of predicting future frames at inference.
These methods either bypass future prediction at test time \citep{yuan2026fastwam,ye2026gigaworld}, distill the model into fewer denoising steps \citep{akbari2026flashwam}, or predict the future in a latent space \citep{luo2026beingh07,chen2026lawam}.
ImageWAM replaces the video model with an image editing model and therefore predicts only a single target frame \citep{zhang2026imagewam}.
Our implementation is based on ImageWAM\@.
OpenWAM systematically examines WAM design choices through controlled experiments \citep{wang2026openwam}.
Its controlled experiments suggest that effective world--action synergy benefits from dedicated action capacity, explicit world-to-action information flow, and synchronized joint denoising.
Our model likewise adopts synchronized joint denoising, but we examine whether dedicated action capacity is in fact necessary.

\paragraph{Actions in the visual domain.}
Several recent methods represent actions within the visual domain, allowing the generative backbone to process them directly.
Action Images renders 7-DoF actions as pixel-grounded, multi-view action videos, thereby enabling the video backbone itself to serve as a policy \citep{zhen2026actionimages}.
Hydra-0 represents actions as pixel motion and employs this shared visual interface across embodiments and video backbones \citep{li2026hydra0}.
It nevertheless relies on a trained action head to map the resulting features to executable actions.
Masked Visual Actions partially reveals a trajectory in the video \citep{alzayer2026maskedvisualactions}.
The same model can thereby predict how the scene responds to an action or infer an action from a desired outcome.
All three methods situate actions in pixel space.
Cosmos Policy instead operates in the latent space of a video model, encoding actions as latent frames \citep{kim2026cosmospolicy}.
Latent action models are also related, as they learn action-like representations from inter-frame changes \citep{chen2025villax,tharwat2025lawm}.
Action-as-Patch likewise represents actions within the visual domain, but without rendering them into pixels.
A parameter-free fixed mapping encodes each action step as a single latent token rather than as an entire latent frame.

\paragraph{Joint modeling across modalities.}
Latent diffusion models denoise images in a compressed latent space \citep{rombach2022ldm}.
Diffusion transformers process this latent as a sequence of patch tokens \citep{peebles2023dit}.
Flow matching is widely adopted for training such models \citep{lipman2022flow}, and recent models further process text and image tokens within a single sequence \citep{esser2024sd3}.
Our backbone, FLUX.2 [klein], belongs to this family \citep{bflklein}.
More broadly, unified multimodal models show that a single model can jointly model multiple modalities once they are represented as tokens.
BAGEL and Emu3.5 are pretrained on interleaved text, image, and video tokens \citep{deng2025bagel,cui2025emu35}.
Cosmos 3 further incorporates action, processing it together with language, image, video, and audio within a single mixture-of-transformers \citep{nvidia2026cosmos3}.
Action-as-Patch exploits the same property within a pretrained image backbone.
Because Action-as-Patch represents actions in the backbone's native token format, our method requires neither a learned action tokenizer nor a dedicated action expert.
When multiple modalities are denoised jointly, they need not share a noise level.
Diffusion Forcing assigns each token an independent noise level \citep{chen2024diffusionforcing}, and several WAMs employ distinct timesteps or schedules for video and action \citep{zhu2025uwm,guo2026xwam,akbari2026flashwam}.
We instead adopt a single noise level for action and future-image tokens.

\section{Method}
\label{sec:method}
\begin{figure}[t]
  \centering
  \includegraphics[width=\textwidth]{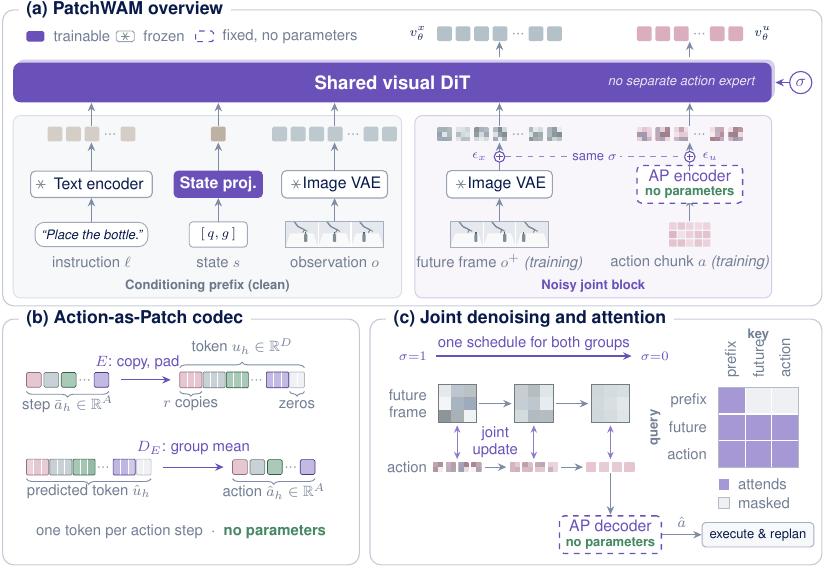}
  \caption{\textbf{Architecture of \model.}
  \textbf{(a)}~Language, robot state, and the current observation form a clean conditioning prefix.
  During training, the future frame and the action chunk are encoded and perturbed with independent noise at a shared noise level~$\sigma$, and one shared visual DiT predicts the velocities of both groups.
  \textbf{(b)}~Action-as-Patch encodes each action step as one token by repeating every component $r=\lfloor D/A\rfloor$ times and padding with zeros, and decodes a token by averaging the copies of each component; the codec has no parameters.
  \textbf{(c)}~At inference, both token groups start from pure noise and are denoised together along one schedule.
  Future-frame and action tokens attend to each other and to the prefix, while the prefix does not attend to the noisy block.
  Token counts and camera views are schematic, and the additions on RoboDojo are not drawn (Section~\ref{sec:formulation}).}
  \label{fig:architecture}
\end{figure}

As discussed in Section~\ref{sec:related-work}, many world action models retain a separate pathway for actions.
The approaches most closely related to ours represent actions within the visual domain, either in pixel space or as entire latent frames.
\model{} instead encodes each action step as a single token in the native token format of the backbone, so that action prediction becomes token prediction within the existing visual path (Figure~\ref{fig:architecture}).
The conditioning prefix combines language features, a projected robot-state token, and the encoded current observation.
During training, the future frame and the action chunk are encoded separately and perturbed with independent Gaussian noise at a shared noise level.
Neither ground-truth target is provided at inference time; instead, both token groups are initialized from Gaussian noise and jointly denoised.
A fixed decoder then recovers the action chunk, which the robot executes before the policy replans.
The model predicts a single future frame rather than a full video, and the actions are obtained without decoding the future-frame latent into RGB\@.

Section~\ref{sec:formulation} formulates the problem and describes the tokenization and the model, and Section~\ref{sec:codec} defines the fixed codec.
Section~\ref{sec:objective} presents the training objective, Section~\ref{sec:attention} describes the token sequence and attention mask, and Section~\ref{sec:inference} details the inference procedure.

\subsection{Problem Setup}
\label{sec:formulation}
We formulate language-conditioned robotic manipulation as the prediction of action chunks.
At each time step $t$, the policy is conditioned on a language instruction $\ell$, a multi-camera observation $o_t$, and a proprioceptive state $s_t$.
It predicts an action chunk of horizon $H$,
\begin{equation}
a_{t:t+H-1}=(a_t,\ldots,a_{t+H-1})\in\mathbb{R}^{H\times A},
\end{equation}
where $A$ denotes the action dimensionality of the robot.
We set $H=16$ in all experiments.
A world action model additionally predicts the future state of the environment \citep{wang2026wamfrontier}.
In \model{}, this future state is represented by the observation at the end of the chunk, $o^{+}=o_{t+H}$, and the model learns the joint conditional distribution
\begin{equation}
p_\theta\big(o^{+},\,a_{t:t+H-1}\;\big|\;\ell,\,o_t,\,s_t\big).
\label{eq:joint}
\end{equation}
The ground-truth future frame is used only as a training target.
At inference time, the future frame and the action chunk are generated jointly, and only the actions are executed.
The subscript $t$ is omitted hereafter when unambiguous.

\paragraph{Tokenization.}
On RoboTwin and LIBERO, the camera views are composed into a single image; on RoboDojo, each view is encoded separately.
A frozen image autoencoder $\mathcal{V}$ converts each input image into a sequence of latent patch tokens,
\begin{equation}
x^{\mathrm{ref}}=\mathcal{V}(o),\qquad x_0=\mathcal{V}(o^{+}),\qquad x^{\mathrm{ref}},\,x_0\in\mathbb{R}^{S\times D},
\end{equation}
where $S$ denotes the number of tokens and $D=128$ denotes the token dimension.
The instruction is embedded by a text encoder $\mathcal{T}$ as $h_\ell=\mathcal{T}(\ell)$, and the proprioceptive state is mapped to a single token $h_s=W_s\,s$ by a trainable linear projection.
Together, these embeddings form the conditioning prefix
\begin{equation}
c=\big[\,h_\ell\,;\;h_s\,;\;x^{\mathrm{ref}}\,\big].
\label{eq:prefix}
\end{equation}
The normalized action chunk is encoded into $H$ tokens of the same dimension, one per step, by the fixed codec described in Section~\ref{sec:codec}, yielding $u_0\in\mathbb{R}^{H\times D}$.
Accordingly, the distribution in Eq.~\eqref{eq:joint} is modeled in token space as $p_\theta(x_0,u_0\mid c)$.

\paragraph{Model.}
A single transformer jointly denoises the future-frame tokens and the action tokens.
Given the noisy tokens $x_\sigma$ and $u_\sigma$ at a shared noise level $\sigma$, it predicts a velocity for each token group:
\begin{equation}
\big(v_\theta^{x},\,v_\theta^{u}\big)=v_\theta\big(x_\sigma,\,u_\sigma,\,\sigma;\;c\big).
\label{eq:model}
\end{equation}
The trainable parameters $\theta$ comprise the transformer weights and $W_s$, together with the LoRA adapters used on RoboDojo.
The image autoencoder and the base weights of the text encoder remain frozen, and the codec is parameter-free.
The model in Eq.~\eqref{eq:model} thus contains no action-specific network, and the gradients of the action loss reach the same transformer that denoises the future frame.
The transformer is initialized from FLUX.2 [klein] 4B Base \citep{bflklein}, and the implementation is based on the ImageWAM codebase \citep{zhang2026imagewam}.
On RoboTwin and LIBERO, $\mathcal{T}$ is instantiated with Qwen3-4B \citep{yang2025qwen3}.
On RoboDojo, $\mathcal{T}$ is Qwen3-VL-4B-Instruct \citep{bai2025qwen3vl} with trainable LoRA adapters \citep{hu2022lora}.
It jointly encodes the instruction, the current head-camera image, and past head-camera frames, and it learns to predict the current subtask as an auxiliary target.
On RoboDojo, the prefix of the transformer also contains latent tokens of past head-camera frames (Section~\ref{sec:setup} and Appendix~\ref{app:implementation}).

\subsection{Action-as-Patch}
\label{sec:codec}
Action-as-Patch maps each action step onto a token whose dimensionality matches that of a visual latent patch, so that actions are presented to the transformer in the same form as image patches.
A fixed encoder--decoder pair implements the mapping on normalized actions.

\paragraph{Normalization.}
Because action dimensions are expressed in heterogeneous physical units, each dimension is first normalized by a dataset-specific transform, $\bar a=N(a)$.
Z-score normalization is adopted for RoboTwin and RoboDojo, and min--max normalization for LIBERO (Section~\ref{sec:setup}).
The codec operates only on normalized values, and the inverse transform $N^{-1}$ is applied after decoding.

\paragraph{Encoder.}
Let $r=\lfloor D/A\rfloor$ denote the replication factor and $\alpha>0$ a fixed scaling constant.
For each step $h$, the encoder replicates every action component $r$ times and zero-pads the remaining $D-rA$ entries:
\begin{equation}
E(\bar a_h)=\alpha\left[
\bar a_{h,1}\mathbf{1}_r^\top,\ldots,
\bar a_{h,A}\mathbf{1}_r^\top,
\mathbf{0}_{D-rA}^\top
\right]^\top.
\label{eq:codec}
\end{equation}
Equivalently, the encoder can be written as $E(\bar a_h)=\alpha P\,\bar a_h$, with the fixed binary matrix
\begin{equation}
P=\begin{bmatrix} I_A\otimes\mathbf{1}_r\\ \mathbf{0}_{(D-rA)\times A}\end{bmatrix}\in\{0,1\}^{D\times A},
\label{eq:codec-matrix}
\end{equation}
where $\otimes$ denotes the Kronecker product.
The encoder is therefore linear, parameter-free, and identical across all steps.
Applying the encoder to each of the $H$ steps of a chunk yields the action tokens introduced in Section~\ref{sec:formulation}, $u_0=\big[E(\bar a_1)^\top;\ldots;E(\bar a_H)^\top\big]\in\mathbb{R}^{H\times D}$.
The replication factor is $r=9$ for the 14-dimensional RoboTwin and RoboDojo actions and $r=18$ for the 7-dimensional LIBERO actions; in both cases, 126 entries carry action values and the remaining two are zero-padded.
The scaling constant is set to $\alpha=1$ throughout.

\paragraph{Decoder.}
The decoder recovers each component by averaging its group of copies:
\begin{equation}
[D_E(u_h)]_i=\frac{1}{\alpha r}
\sum_{j=(i-1)r+1}^{ir}u_{h,j},\qquad i=1,\ldots,A.
\label{eq:decode}
\end{equation}
Because $P^\top P=rI_A$, this readout coincides with the Moore--Penrose pseudoinverse of the encoder, $D_E(u_h)=(\alpha P)^{+}u_h$.
This identity yields two properties.
First, the decoder is an exact left inverse of the encoder, $D_E(E(\bar a_h))=\bar a_h$ up to floating-point error, and the codec therefore requires no reconstruction loss.
Second, for a predicted token $\hat u_h$ lying outside the range of $E$, the decoder returns the action whose encoding is nearest in the least-squares sense:
\begin{equation}
D_E(\hat u_h)=\operatorname*{arg\,min}_{\bar a\in\mathbb{R}^{A}}\big\|E(\bar a)-\hat u_h\big\|_2^2 .
\label{eq:lsq}
\end{equation}
The readout is thus an orthogonal projection onto the range of the encoder followed by exact inversion, not a learned prediction head.
Actions are never passed through the image autoencoder or quantized into a discrete vocabulary.

\paragraph{Actions as patches.}
An action token shares the dimensionality $D$ of a visual latent patch.
It is therefore processed by the same learned input and output projections as a visual patch, and no token-type embedding is added.
The two token groups are distinguished structurally by their positional coordinates and sequence locations (Section~\ref{sec:attention}); they also enter separate loss terms (Section~\ref{sec:objective}).
Replication further distributes each component across $r$ entries.
Since the entries of the action noise $\epsilon_u$ (Section~\ref{sec:objective}) are independent standard Gaussian variables, the mean of the $r$ noisy copies of a component at noise level $\sigma$ has a noise variance of $\sigma^2/r$.
This redundancy therefore raises the signal-to-noise ratio of each component in the noisy input by a factor of $r$ relative to a single copy.
The same reasoning does not extend to the output: because the predicted copies of a component are not independent, averaging them does not necessarily reduce the prediction error by a fixed factor.

\paragraph{Comparison designs.}
The fixed codec occupies one end of a spectrum of designs.
At the other end, the dual-expert control augments the same backbone with a separate action expert (Figure~\ref{fig:intro-overview}, left); other factors changed by this comparison are discussed in Section~\ref{sec:sparse}.
An intermediate learned-linear control replaces $E$ and $D_E$ with trainable linear maps while holding the shared transformer, training data, and optimization budget fixed (Section~\ref{sec:mechanisms}).
The first comparison tests whether a full action expert is necessary, and the second whether even a lightweight learned mapping is required.

\subsection{Training Objective}
\label{sec:objective}
Training starts from the clean future-frame tokens $x_0$ and action tokens $u_0$ of Section~\ref{sec:formulation}.
For each training sample, we draw a single noise level $\sigma$ and two independent standard Gaussian tensors $\epsilon_x$ and $\epsilon_u$:
\begin{equation}
x_\sigma=(1-\sigma)x_0+\sigma\epsilon_x,\qquad
u_\sigma=(1-\sigma)u_0+\sigma\epsilon_u.
\end{equation}
The two groups thus share the noise level but not the noise realization.
The noise level is obtained by shifting a uniform variable,
\begin{equation}
\sigma=\phi_s(\tau)=\frac{s\,\tau}{1+(s-1)\,\tau},\qquad \tau\sim\mathcal{U}(0,1),
\label{eq:shift}
\end{equation}
with $s=5$, following the schedule shift introduced for rectified-flow transformers \citep{esser2024sd3}.
Since $\phi_s(\tau)\geq\tau$ for $s>1$, the shift concentrates training on high noise levels: only one sixth of the sampled noise levels fall below $\sigma=\tfrac{1}{2}$.
The transformer of Eq.~\eqref{eq:model} receives the prefix followed by the noisy tokens $[x_\sigma;u_\sigma]$ and predicts $v_\theta^x$ and $v_\theta^u$, whose targets are $\epsilon_x-x_0$ and $\epsilon_u-u_0$.
The objective is
\begin{align}
\mathcal{L}=\mathbb{E}\big[
w(\sigma)\big(&\lambda_x\,\operatorname{MSE}(v_\theta^x,\epsilon_x-x_0)
\nonumber\\
&+\lambda_u\,\operatorname{MSE}_{\mathrm{valid}}(v_\theta^u,\epsilon_u-u_0)\big)\big],
\label{eq:loss}
\end{align}
where $\lambda_x=0.5$ and $\lambda_u=1.0$ weight the two modalities, and $w(\sigma)$ weights the noise level,
\begin{equation}
w(\sigma)=\frac{1}{Z}\Big(\exp\!\big(-2\,(\sigma-\tfrac{1}{2})^{2}\big)-e^{-1/2}\Big).
\label{eq:weight}
\end{equation}
This weight peaks at $\sigma=\tfrac{1}{2}$ and vanishes at $\sigma=0$ and $\sigma=1$, and the constant $Z$ normalizes its expectation under Eq.~\eqref{eq:shift} to one.
The shift thus sets how often each noise level is sampled, and the weight sets how much each sample contributes to the loss.
Each term is averaged over the tokens of its group before the weighted sum, so that $\lambda_x$ and $\lambda_u$, rather than the token counts, set the relative weight of the two terms.
The action term, $\operatorname{MSE}_{\mathrm{valid}}$, averages only over the action steps that lie within the demonstration and excludes the padded steps of chunks that extend past its end.
The two zero-padded entries of each action token are retained in the regression and discarded only at decoding.
On RoboDojo, the cross-entropy loss of the subtask prediction is added to $\mathcal{L}$ (Appendix~\ref{app:implementation}).

\subsection{Token Sequence and Attention}
\label{sec:attention}
The input sequence begins with the clean conditioning prefix $c$ defined in Eq.~\eqref{eq:prefix}.
The prefix is followed by the noisy joint block, in which the future-frame tokens $x_\sigma$ precede the action tokens $u_\sigma$.
Prefix tokens attend only to prefix tokens.
Future-frame and action tokens attend to the prefix and to all tokens of the noisy block.
The token sequence and the attention mask for a RoboTwin sample are shown in Figure~\ref{fig:attention-case} of Appendix~\ref{app:attention}.

Each token carries a positional coordinate with four axes, and these coordinates distinguish the token groups.
On the first axis, current-frame, future-frame, and action tokens take the values 10, 0, and 20, respectively; image tokens carry their row and column on the second and third axes, and text and action tokens carry their sequence and step indices on the fourth.
The values 10, 0, and 20 only separate the groups and do not represent physical time.
On RoboDojo, the additional camera views and the past frames take further values on the first axis (Appendix~\ref{app:implementation}).

\subsection{Inference}
\label{sec:inference}
At inference time, the model encodes the current observation and initializes the joint latent $z_0=[x_1;u_1]$ with standard Gaussian noise, which corresponds to $\sigma_0=1$.
It then integrates the predicted velocity with the Euler method along the decreasing schedule $\sigma_k=\phi_s(1-k/K)$, $k=0,\ldots,K$, which uses the same shift as in training:
\begin{equation}
z_{k+1}=z_k+(\sigma_{k+1}-\sigma_k)\,v_\theta(z_k,\sigma_k;\,c).
\end{equation}
The default is $K=20$ solver steps; all evaluations in this paper use $K=10$ (Section~\ref{sec:setup}).
The final action tokens are decoded with Eq.~\eqref{eq:decode} and mapped back to physical units with $N^{-1}$.
The robot executes the chunk, or its first part, after which the policy replans from the new observation (Appendix~\ref{app:implementation}).
The future-frame latent need not be decoded into RGB\@.

Each solver step updates the future-frame and action tokens jointly.
A dual-expert model, by contrast, can compute the keys and values of its visual transformer once and then denoise only the actions; Section~\ref{sec:cost} discusses the resulting difference in inference cost.
The architecture permits information exchange between the two groups during denoising.
Section~\ref{sec:mechanisms} tests whether the trained model uses this exchange with the isolated-attention control, which blocks attention between the two groups.

\section{Experiments}
\label{sec:experiments}
We first describe the experimental setup (Section~\ref{sec:setup}) and then present the main results on RoboTwin~2.0 (Section~\ref{sec:robotwin-main}).
Section~\ref{sec:sparse} compares \model{} with a dual-expert control under matched training conditions, and Section~\ref{sec:benchmarks} reports results on additional benchmarks.

\subsection{Experimental Setup}
\label{sec:setup}
\paragraph{Benchmarks and training data.}
RoboTwin~2.0 \citep{chen2025robotwin2} contains 50 bimanual manipulation tasks, each evaluated in clean and randomized scenes.
We use two training protocols on this benchmark.
In the full-data protocol, training uses 50 clean and 500 randomized demonstrations per task, and the strongest \model{} configuration adds 2,100 augmented demonstrations for seven tasks with low success rates.
This protocol includes randomized demonstrations and therefore does not measure generalization from clean to randomized scenes.
It also differs from the fixed-data leaderboard protocol, which uses 50 clean demonstrations per task \citep{robotwinleaderboard2026}.
In the clean-to-random (C2R) protocol, training uses only these 50 clean demonstrations per task, as on the leaderboard, and the success rate in randomized scenes is the main measure of generalization.
The best C2R configuration applies photometric, AdaIN, and FFT-based augmentation, and we report its checkpoint after 140k updates.
Online augmentation of clean demonstrations is not equivalent to training on additional randomized demonstrations.
LIBERO \citep{liu2023libero} comprises four suites of ten tasks each.
LIBERO-Plus \citep{fei2026liberoplus} perturbs these tasks, and we report its success rates separately for training on the original LIBERO data and for training on augmented LIBERO-Plus data.
RoboDojo \citep{chen2026robodojo} spans 42 simulated tasks and five capability dimensions.
Its generalization dimension is evaluated under standard and randomized conditions.
We evaluate \model{} on all tasks and conditions of the benchmark.

\paragraph{Matched dual-expert control.}
The dual-expert control adds a separate action expert with 0.64B trainable parameters to the same backbone (Section~\ref{sec:codec} and Appendix~\ref{app:implementation}).
\model{} and the control are trained on the clean and randomized demonstrations of the full-data protocol, without the augmented demonstrations, and share all optimization settings.
Both runs retain one of every 20 possible training-window start positions, which reduces the number of training windows rather than the number of demonstrations.

\paragraph{Implementation.}
For RoboTwin, the three camera views are composed into a $288\times256$ image, and both actions and proprioceptive states are 14-dimensional.
For LIBERO, two camera views are composed horizontally into a $224\times448$ image, actions are 7-dimensional, and proprioceptive states are 8-dimensional.
For RoboDojo, the three camera views remain separate $256\times256$ images, and both actions and proprioceptive states are 14-dimensional.
On RoboDojo, the prefix also includes up to 20 past head-camera frames at one-second intervals, and the LoRA adapters of the text encoder add 132.1M trainable parameters.
Actions are z-score normalized on RoboTwin and RoboDojo and min--max normalized on LIBERO\@.
The global batch size is 256, except that it is 128 for LIBERO and 64 for the matched comparison.
Appendix~\ref{app:implementation} lists further run-specific settings.

\paragraph{Evaluation protocol.}
Success rates (SR) are reported in percent.
On RoboTwin~2.0, the full-data and C2R evaluations run 100 episodes per task and scene condition, or 10,000 episodes in total.
Both runs of the matched comparison are evaluated with 10 episodes per task and scene condition, or 1,000 episodes per checkpoint.
RoboDojo results also include the Score, the metric by which the leaderboard ranks its entries \citep{robodojoleaderboard2026}.
All evaluations use $K=10$ solver steps instead of the default $K=20$ (Section~\ref{sec:inference}).
Each result of \model{} comes from a single evaluation run, so no standard deviations are reported.
The baselines differ in demonstration counts, augmentation, and checkpoint selection, so benchmark comparisons do not isolate the effect of the architecture.
Section~\ref{sec:sparse} examines this effect under matched training conditions.

\subsection{Main Results on RoboTwin 2.0}
\label{sec:robotwin-main}
RoboTwin~2.0 is the primary benchmark in our evaluation.
Tables~\ref{tab:robotwin-full} and~\ref{tab:robotwin-c2r} compare \model{} with prior methods under the two training protocols of Section~\ref{sec:setup}, and the C2R baselines are taken from the official leaderboard \citep{robotwinleaderboard2026}.
\begin{table}[t]
\centering
\caption{\textbf{RoboTwin 2.0 success rate (\%) under the full-data protocol:} training on clean and randomized demonstrations and evaluation in clean and randomized scenes. Best in \textbf{bold}; second-best \underline{underlined}.}
\label{tab:robotwin-full}
\footnotesize
\setlength{\tabcolsep}{5pt}
\begin{tabular}{lccc}
\toprule
\textbf{Method} & \textbf{Clean} & \textbf{Randomized} & \textbf{Average} \\
\midrule
\multicolumn{4}{c}{\textbf{VLA}} \\
\midrule
ZR-0~\citep{li2026zr0} & 88.70 & 87.98 & 88.34 \\
HoloBrain-0~\citep{lin2026holobrain0} & 91.30 & 90.80 & 91.05 \\
HoloBrain-0-QW~\citep{lin2026holobrain0} & 91.90 & 92.30 & 92.10 \\
LingBot-VA~\citep{li2026lingbotva} & 92.93 & 91.55 & 92.24 \\
AIM~\citep{fan2026aim} & \underline{94.00} & 92.10 & 93.05 \\
G0.5~\citep{liu2026g05} & 93.70 & 92.80 & 93.25 \\
LingBot-VA 2.0~\citep{zhang2026lingbotva2} & 93.80 & 93.40 & 93.60 \\
Next Forcing~\citep{xu2026nextforcing} & \textbf{94.10} & \underline{93.50} & \underline{93.80} \\
Qwen-RobotManip~\citep{yuan2026qwenrobotmanip} & 93.70 & \textbf{94.00} & \textbf{93.85} \\
\midrule
\multicolumn{4}{c}{\textbf{WAM}} \\
\midrule
X-WAM~\citep{guo2026xwam} & 89.80 & 90.70 & 90.25 \\
LaWAM~\citep{chen2026lawam} & 92.64 & 89.80 & 91.22 \\
Faster-WAM~\citep{zhao2026fasterwam} & 92.80 & 92.30 & 92.55 \\
SelfWAM~\citep{pan2026selfwam} & 92.16 & 93.08 & 92.62 \\
MECo-WAM~\citep{zhang2026mecowam} & 93.26 & 91.98 & 92.62 \\
ImageWAM~\citep{zhang2026imagewam} & 93.20 & 93.56 & 93.38 \\
ABot-M0.5~\citep{chen2026abotm05} & 94.00 & 94.20 & 94.10 \\
MotuBrain~\citep{motubrain2026} & \underline{95.80} & \underline{96.08} & \underline{95.94} \\
\textbf{\model{} (Ours)} & \textbf{96.04} & \textbf{96.20} & \textbf{96.12} \\
\bottomrule
\end{tabular}
\end{table}

\begin{table}[t]
\centering
\caption{\textbf{RoboTwin 2.0 success rate (\%) under the clean-to-random protocol:} training on clean demonstrations only and evaluation in clean and randomized scenes. Best in \textbf{bold}; second-best \underline{underlined}.}
\label{tab:robotwin-c2r}
\footnotesize
\setlength{\tabcolsep}{5pt}
\begin{tabular}{lccc}
\toprule
\textbf{Method} & \textbf{Clean} & \textbf{Randomized} & \textbf{Average} \\
\midrule
\multicolumn{4}{c}{\textbf{VLA}} \\
\midrule
StarVLA~\citep{starvla2026starvla} & 46.52 & 3.16 & 24.84 \\
GalaxeaVLA~\citep{jiang2025g0} & 62.70 & 12.72 & 37.71 \\
Xiaomi Robotics-0~\citep{cai2026xiaomirobotics0} & 62.90 & 18.20 & 40.55 \\
EventVLA~\citep{yang2026eventvla} & 65.60 & 15.70 & 40.65 \\
Spatial Forcing~\citep{li2026spatialforcing} & \underline{77.20} & 26.74 & 51.97 \\
OLA-Geo~\citep{robotwinleaderboard2026} & \textbf{82.96} & 33.64 & 58.30 \\
$\pi_{0.5}$~\citep{intelligence2025pi05} & 70.70 & 46.00 & 58.35 \\
GigaBrain-0.7~\citep{gigabrain2026gigabrain07} & 66.80 & \textbf{67.90} & \underline{67.35} \\
OLA-Sem~\citep{robotwinleaderboard2026} & 75.12 & \underline{67.56} & \textbf{71.34} \\
\midrule
\multicolumn{4}{c}{\textbf{WAM}} \\
\midrule
AHA-WAM~\citep{cai2026ahawam} & 64.30 & 3.20 & 33.75 \\
FastWAM~\citep{yuan2026fastwam} & 77.80 & 1.90 & 39.85 \\
ABot-M0~\citep{yang2026abotm0} & 57.40 & 30.36 & 43.88 \\
X-WAM~\citep{guo2026xwam} & 70.00 & 25.80 & 47.90 \\
4D-WAM~\citep{yang20264dwam} & \underline{81.50} & \underline{41.80} & \underline{61.65} \\
\textbf{\model{} (Ours)} & \textbf{91.56} & \textbf{66.72} & \textbf{79.14} \\
\bottomrule
\end{tabular}
\end{table}

\paragraph{Full-data protocol.}
Under the full-data protocol, \model{} achieves 96.04\% success in clean scenes and 96.20\% in randomized scenes after 110k updates, for an average of 96.12\% (Table~\ref{tab:robotwin-full}).
Both success rates are the highest among the listed methods, and the two scene conditions differ by only 0.16 percentage points.
The strongest listed baseline, MotuBrain, reaches an average of 95.94\%, 0.18 percentage points below \model{}.
ImageWAM reaches 93.38\%; it uses an image-editing backbone with a separate action expert and provides the codebase on which \model{} is built.
The compared methods are not matched in training data, and the \model{} configuration includes 2,100 augmented demonstrations, so these differences do not isolate the effect of the architecture.
Section~\ref{sec:sparse} examines this effect under matched conditions.
Appendix~\ref{app:training-curve} traces the success rate over a full-data training run without the augmented demonstrations.

\paragraph{Clean-to-random generalization.}
Under the C2R protocol, \model{} achieves 66.72\% success in randomized scenes and 91.56\% in clean scenes, for an average of 79.14\% (Table~\ref{tab:robotwin-c2r}).
This average is the highest among the listed methods, while the randomized-scene success rate remains below the strongest listed result of 67.90\%.
Only GigaBrain-0.7 and OLA-Sem succeed more often in randomized scenes, by 1.18 and 0.84 percentage points, whereas \model{} exceeds their clean-scene success rates by 24.76 and 16.44 points.
The WAM SoTA under this protocol, 4D-WAM, reaches 41.80\% in randomized scenes.
The drop from clean to randomized scenes is 24.84 percentage points for \model{}, whereas 4D-WAM, X-WAM, FastWAM, and AHA-WAM lose between 39.70 and 75.90 points.
Under the full-data protocol, which includes randomized demonstrations, the two scene conditions differ by only 0.16 percentage points.
This contrast is consistent with the hypothesis that the C2R drop arises mainly from the absence of randomized demonstrations in training.
Because the C2R configuration applies photometric, AdaIN, and FFT-based augmentation, the smaller drop of \model{} relative to the other listed WAMs cannot be attributed to the architecture alone.
Appendix~\ref{app:c2r-per-task} breaks these results down by task, and Appendix~\ref{app:c2r-qualitative} shows example rollouts.

\subsection{Does Action Prediction Need a Dedicated Path?}
\label{sec:sparse}
The matched comparison tests whether a policy remains effective when the dedicated action expert is removed and actions are represented directly as tokens of the visual backbone.
\begin{table}[htbp]
\centering
\caption{\textbf{Effect of a dedicated action expert on RoboTwin~2.0.} Success rates (percent) of the matched comparison after 41,940 updates, with one of every 20 window starts retained and 10 episodes per task and condition.}
\label{tab:sparse}
\small
\begin{tabular}{cccc}
\toprule
\textbf{Action expert} & \textbf{Clean} & \textbf{Randomized} & \textbf{Average}\\
\midrule
\cmark{} & 77.2 & 79.6 & 78.4\\
\xmark{} & \best{87.6} & \best{88.4} & \best{88.0}\\
\midrule
$\Delta$ & +10.4 & +8.8 & +9.6\\
\bottomrule
\end{tabular}
\end{table}
After 41,940 updates under matched training data and optimization, \model{} reaches an average success rate of 88.0\%, compared with 78.4\% for the dual-expert control (Table~\ref{tab:sparse}).
The comparison shows that, in this setting, a dedicated action expert is not needed for a high success rate.
However, the two models also differ in token interaction and parameterization, so the gain cannot be attributed to a single factor.
The comparison is limited to this window-sampling regime and does not imply a twenty-fold reduction in demonstration collection.

\begin{figure}[htbp]
  \centering
  \includegraphics[width=\textwidth]{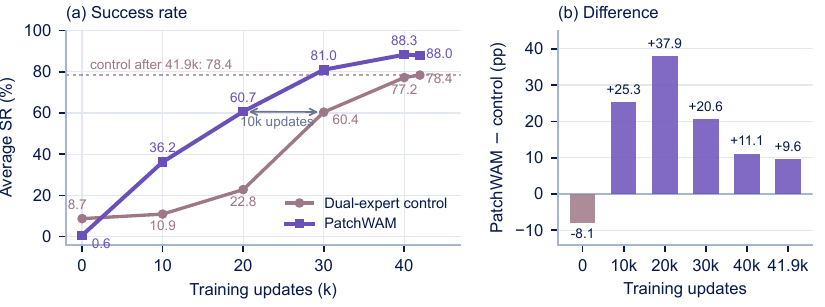}
  \caption{Learning curves of the matched comparison on RoboTwin~2.0. \textbf{(a)}~Average success rate of \model{} and the dual-expert control at intermediate checkpoints of the two runs; the first point of each curve is the evaluation before training. \textbf{(b)}~Difference between the two success rates (\model{} minus control).}
  \label{fig:matched-curve}
\end{figure}

\paragraph{Learning speed.}
Intermediate checkpoints of the same two runs show that \model{} also learns faster (Figure~\ref{fig:matched-curve}).
Before training, the dual-expert control already succeeds in 8.7\% of the episodes and \model{} in 0.6\%.
Nine tasks have success conditions that can be triggered by a single contact or a short motion: pressing the stapler, clicking the alarm clock and the bell, turning the switch, opening the microwave and the laptop, moving the playing card away, and the two bottle-shaking tasks.
These tasks account for 7.8 of the 8.7 percentage points, because the untrained action expert occasionally satisfies their success conditions by chance.
After 20k updates, \model{} reaches 60.7\%, whereas the dual-expert control needs 30k updates to reach a comparable 60.4\%.
After 30k updates, \model{} reaches 81.0\% and thus exceeds the final success rate of the control, 78.4\% after 41,940 updates.
The difference is largest at 20k updates, where it amounts to 37.9 percentage points, and narrows to 9.6 points by the end of training.
Because both runs use the same data and batch size, fewer updates correspond directly to fewer training samples.
Each curve comes from a single run evaluated with 10 episodes per task and condition, so differences between individual checkpoints are subject to sampling error.

\subsection{Additional Benchmarks}
\label{sec:benchmarks}
\begin{figure}[htbp]
  \centering
  \includegraphics[width=0.48\textwidth]{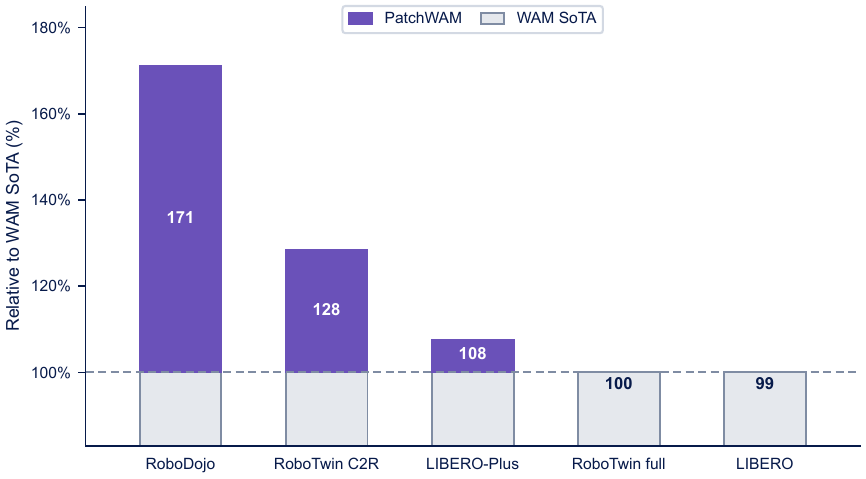}
  \caption{Success rate of \model{} relative to the WAM SoTA on five benchmark settings (relative score on RoboDojo). The WAM SoTA is MotuBrain on RoboTwin full-data, 4D-WAM on RoboTwin C2R, ABot-M0.5 on LIBERO, ImageWAM on LIBERO-Plus, and OpenWAM-$\alpha$ on RoboDojo.}
  \label{fig:benchmark-summary}
\end{figure}

Figure~\ref{fig:benchmark-summary} compares \model{} with the WAM state of the art (WAM SoTA) on each of the five settings.
Tables~\ref{tab:libero-results},~\ref{tab:libero-plus-results}, and~\ref{tab:robodojo-comparison} list the full comparisons on LIBERO, LIBERO-Plus, and RoboDojo, with RoboDojo baselines taken from the official leaderboard \citep{robodojoleaderboard2026}.

\paragraph{LIBERO.}
On LIBERO, \model{} achieves an average success rate of 98.4\%, whereas the strongest listed baseline, QuoVLA, reaches 99.6\% (Table~\ref{tab:libero-results}).
Across the four suites, \model{} reaches 96.5\% on Spatial, 100.0\% on Object, 99.0\% on Goal, and 98.0\% on Long.
Long is the weakest suite for 13 of the 14 listed methods, whereas \model{} succeeds more often on Long than on Spatial.
Only three listed methods exceed its 98.0\% on Long.
On Object, \model{} matches the best listed result of 100.0\%.
Its average is limited mainly by the Spatial suite, on which 11 listed methods succeed more often.
Because three listed methods already exceed 99\% on average, the benchmark appears close to saturation, and the remaining differences are small.
Appendix~\ref{app:additional} reports results on LIBERO-Pro.
\begin{table}[!htbp]
\centering
\caption{\textbf{LIBERO success rate (\%) on the four task suites.} Best in \textbf{bold}.}
\label{tab:libero-results}
\footnotesize
\setlength{\tabcolsep}{4pt}
\begin{tabular}{lccccc}
\toprule
\textbf{Method} & \textbf{Spatial} & \textbf{Object} & \textbf{Goal} & \textbf{Long} & \textbf{Average} \\
\midrule
\multicolumn{6}{c}{\textbf{VLA}} \\
\midrule
OpenVLA~\citep{kim2024openvla} & 84.7 & 88.4 & 79.2 & 53.7 & 76.5 \\
WorldVLA~\citep{cen2025worldvla} & 87.6 & 96.2 & 83.4 & 60.0 & 81.8 \\
UniVLA~\citep{wang2026univla} & 95.4 & 98.8 & 93.6 & 94.0 & 95.4 \\
F1-VLA~\citep{lv2025f1vla} & 98.2 & 97.8 & 95.4 & 91.3 & 95.7 \\
OpenVLA-OFT~\citep{kim2025openvlaoft} & 97.6 & 98.4 & 97.9 & 94.5 & 97.1 \\
TurboVLA~\citep{xie2026turbovla} & 99.2 & 99.8 & 97.4 & 94.2 & 97.6 \\
StarVLA-$\alpha$ (Specialist)~\citep{ye2026starvlaalpha} & 98.7 & 99.7 & 98.6 & 94.2 & 97.8 \\
X-VLA~\citep{zheng2026xvla} & 98.2 & 98.6 & 97.8 & 97.6 & 98.1 \\
S$^2$-VLA~\citep{xie2026s2vla} & 98.4 & 99.6 & 98.4 & 96.4 & 98.2 \\
FocusVLA~\citep{zhang2026focusvla} & 99.6 & \best{100.0} & 98.8 & 96.2 & 98.6 \\
GeoAlign~\citep{chen2026geoalign} & \best{100.0} & 99.6 & \underline{99.8} & 96.6 & 99.0 \\
CORAL (SimVLA)~\citep{luo2026coral,luo2026simvla} & 99.6 & 99.8 & 99.0 & \best{98.8} & \underline{99.3} \\
QuoVLA~\citep{wang2026quovla} & \underline{99.8} & \underline{99.9} & \best{100.0} & \underline{98.7} & \best{99.6} \\
\midrule
\multicolumn{6}{c}{\textbf{WAM}} \\
\midrule
ABot-M0.5~\citep{chen2026abotm05} & \best{100.0} & \underline{99.8} & \best{99.4} & \textbf{98.4} & \best{99.4} \\
\model{} (Ours) & \underline{96.5} & \best{100.0} & \underline{99.0} & \underline{98.0} & \underline{98.4} \\
\bottomrule
\end{tabular}
\end{table}

\FloatBarrier

\paragraph{LIBERO-Plus.}
On LIBERO-Plus, training on the original LIBERO data yields a total success rate of 80.2\%, and training on augmented LIBERO-Plus data raises it to 91.8\% after 75k updates; Table~\ref{tab:libero-plus-results} lists the second result.
The latter result reflects training on perturbed data rather than zero-shot robustness.
This total is the highest among the listed methods, 0.41 percentage points above Hermite-VLAReg, and \model{} also achieves the highest success rates under camera and sensor-noise perturbations, at 97.9\% each.
ImageWAM, which uses a 9B FLUX.2 backbone instead of the 4B backbone of \model{}, reaches 85.3\%.
The largest differences from ImageWAM arise under camera and robot initial-state perturbations, at 18.1 and 12.3 percentage points, while ImageWAM succeeds more often under language perturbations, at 95.2\% compared with 89.1\%.
The weakest category of \model{} is robot initial-state perturbation, at 71.0\%; four listed methods succeed more often in this category, led by the zero-shot QuoVLA result of 87.6\%.
These results indicate that robustness to visual perturbations does not necessarily carry over to variations in the initial robot state.
\begin{table}[t]
\centering
\caption{\textbf{LIBERO-Plus success rate by perturbation category.} Best in \textbf{bold} and second-best \underline{underlined} within each group.}
\label{tab:libero-plus-results}
\footnotesize
\setlength{\tabcolsep}{4pt}
\resizebox{\linewidth}{!}{%
\begin{tabular}{lcccccccc}
\toprule
\textbf{Method} & \textbf{Camera} & \textbf{Robot} & \textbf{Language} & \textbf{Light} & \textbf{Background} & \textbf{Noise} & \textbf{Layout} & \textbf{Total} \\
\midrule
\multicolumn{9}{c}{\textbf{VLA}} \\
\midrule
$\pi_0$~\citep{black2024pi0} & 79.6 & 21.1 & 72.5 & 84.7 & 86.2 & 68.3 & 69.4 & 68.8 \\
OpenVLA*-Full~\citep{kim2024openvla} & 69.4 & 49.6 & 66.3 & 88.2 & 88.5 & 78.7 & 70.3 & 73.0 \\
$\pi_{0.5}$~\citep{intelligence2025pi05} & 70.3 & 41.7 & 81.1 & 97.3 & 94.6 & 71.8 & 84.9 & 77.4 \\
InternVL-3.5 + DiT~\citep{wang2025internvl35} & 94.2 & 32.4 & 64.5 & 94.7 & 93.0 & 94.7 & 74.8 & 78.3 \\
ELAN4D ($\pi_{0.5}$)~\citep{he2026elan4d} & 63.7 & 70.7 & 77.8 & 89.8 & 91.4 & 79.9 & 81.4 & 79.2 \\
GR00T-N1.6~\citep{nvidia2025gr00tn16} & 92.6 & 33.5 & 80.1 & 93.6 & 95.4 & 93.6 & 75.0 & 80.5 \\
OpenVLA-OFT+~\citep{kim2025openvlaoft} & 92.8 & 30.3 & 85.8 & 94.9 & 93.9 & 89.3 & 77.6 & 80.7 \\
SRPO~\citep{fei2026srpo} & 83.4 & 62.0 & 73.6 & 97.2 & 97.7 & 85.7 & 75.2 & 82.1 \\
RoVLA~\citep{luo2026rovla} & \textbf{96.6} & 32.0 & \textbf{91.5} & 95.9 & 96.1 & 95.1 & 74.1 & 83.0 \\
$\pi_{0.5}$ + AXIS~\citep{zhao2026axis} & 83.8 & 78.2 & \underline{88.3} & 96.5 & \underline{98.1} & \underline{96.2} & 85.5 & 89.5 \\
CAC-VLA~\citep{xiong2026cacvla} & 91.2 & 78.4 & 83.3 & 97.5 & 97.1 & 95.4 & 87.8 & 90.1 \\
QuoVLA~\citep{wang2026quovla} & 82.3 & \textbf{87.6} & 88.2 & \underline{98.2} & 95.9 & 90.8 & 89.3 & 90.3 \\
Anchor-Align VLA~\citep{dalal2026anchoralign} & \underline{96.3} & 59.1 & 87.2 & \textbf{99.0} & \textbf{99.6} & \textbf{96.9} & \textbf{97.4} & \underline{90.8} \\
Hermite-VLAReg~\citep{lv2026hermitevla} & 89.2 & \underline{85.4} & 87.3 & 97.7 & 96.7 & 93.8 & \underline{89.9} & \textbf{91.4} \\
\midrule
\multicolumn{9}{c}{\textbf{WAM}} \\
\midrule
Fast-WAM~\citep{yuan2026fastwam} & 16.4 & 44.5 & 68.9 & 78.2 & 53.7 & 37.7 & 60.7 & 50.0 \\
Being-H0.7~\citep{luo2026beingh07} & - & - & - & - & - & - & - & 82.1 \\
Cosmos-Policy~\citep{kim2026cosmospolicy} & 75.8 & 63.3 & 81.7 & \underline{96.5} & 88.9 & 92.7 & 82.2 & 82.2 \\
ImageWAM~\citep{zhang2026imagewam} & \underline{79.8} & 58.7 & \textbf{95.2} & 96.1 & \underline{91.2} & \underline{93.3} & \underline{83.1} & \underline{85.3} \\
ABot-M0.5~\citep{chen2026abotm05} & 70.5 & \textbf{87.4} & 88.6 & 94.0 & 89.7 & 75.5 & 85.2 & 83.4 \\
\textbf{\model{}} & \textbf{97.9} & \underline{71.0} & \underline{89.1} & \textbf{98.5} & \textbf{96.9} & \textbf{97.9} & \textbf{92.3} & \textbf{91.8} \\
\bottomrule
\end{tabular}%
}
\end{table}

\FloatBarrier

\paragraph{RoboDojo.}
On RoboDojo, \model{} obtains a Score of 29.39 and an average success rate of 23.60\% (Table~\ref{tab:robodojo-comparison}).
The Score assigns 100 points to a successful episode and partial credit to an unsuccessful episode according to the intermediate steps completed.
Both the Score and the success rate are averaged over the tasks of each capability dimension and then over the five dimensions with equal weight, so the standard and randomized conditions of the generalization dimension each contribute one tenth \citep{chen2026robodojo}.
Both are the highest among the entries of the leaderboard on September 15, 2026; the next entry, DM0.5, has a Score of 24.90 and a success rate of 19.34\%.\footnote{As of September 22, 2026, two entries added to the leaderboard after September 15 exceed at least one of these values: Liber-0 Preview has a Score of 30.74 and a success rate of 25.52\%, and Liber-0 Lite has a success rate of 24.23\% at a Score of 29.24.}
The WAM SoTA on the leaderboard, OpenWAM-$\alpha$, has a Score of 17.18.
Performance varies strongly across capability dimensions.
\model{} reaches the highest success rate among the listed methods under both generalization conditions, on precise manipulation, and in open-vocabulary instruction following.
Under the standard and randomized generalization conditions, it succeeds in 33.33\% and 16.33\% of the episodes, compared with at most 28.00\% and 6.00\% for the other listed methods.
It reaches 23.00\% on precise manipulation, compared with at most 20.42\%, and 22.00\% in open-vocabulary instruction following, compared with at most 4.25\%.
On long-horizon tasks, it reaches 31.50\%, below the 32.25\% of G0.5, and on memory tasks it reaches 16.67\%, compared with 47.44\% for DM0.5.
\begin{table}[t]
\centering
\caption{\textbf{RoboDojo simulation results across capability dimensions}, with the score and the success rate (SR, percent) of each dimension.}
\label{tab:robodojo-comparison}
\footnotesize
\setlength{\tabcolsep}{3pt}
\resizebox{\linewidth}{!}{%
\begin{tabular}{lcccccccccccccc}
\toprule
 & \multicolumn{2}{c}{\textbf{Gen-Std}} & \multicolumn{2}{c}{\textbf{Gen-Rand}} & \multicolumn{2}{c}{\textbf{Precision}} & \multicolumn{2}{c}{\textbf{Long}} & \multicolumn{2}{c}{\textbf{Memory}} & \multicolumn{2}{c}{\textbf{Open}} & \multicolumn{2}{c}{\textbf{Average}} \\
\cmidrule(lr){2-3}\cmidrule(lr){4-5}\cmidrule(lr){6-7}\cmidrule(lr){8-9}\cmidrule(lr){10-11}\cmidrule(lr){12-13}\cmidrule(lr){14-15}
\textbf{Method} & Score & SR & Score & SR & Score & SR & Score & SR & Score & SR & Score & SR & Score & SR \\
\midrule
\multicolumn{15}{c}{\textbf{VLA}} \\
\midrule
StarVLA-PI\_v3~\citep{starvla2026starvla} & 19.08 & 14.44 & 3.36 & 1.67 & 17.77 & 12.50 & 18.46 & 11.00 & 4.59 & 4.00 & 2.03 & 2.00 & 10.81 & 7.51 \\
InternVLA-A1.5~\citep{ma2026internvlaa15} & 16.81 & 11.78 & 3.90 & 1.89 & 15.23 & 10.17 & 23.80 & 13.75 & 4.93 & 3.56 & 1.43 & 1.42 & 11.15 & 7.14 \\
$\pi_{0.5}$~\citep{intelligence2025pi05} & 20.93 & 14.89 & 5.82 & 1.44 & 12.40 & 5.50 & 23.54 & 14.67 & 5.78 & 4.56 & 1.98 & 1.67 & 11.41 & 6.91 \\
Spatial Forcing~\citep{li2026spatialforcing} & 21.25 & 14.89 & 6.98 & 3.78 & 17.32 & 10.58 & 23.26 & 14.58 & 5.43 & 4.11 & 1.78 & 1.58 & 12.38 & 8.04 \\
Hy-Embodied-0.5-VLA~\citep{zhang2026hyembodied05vla} & 21.98 & 16.56 & 1.57 & 0.22 & 13.81 & 8.00 & 25.74 & 14.92 & 13.37 & 12.11 & 0.65 & 0.58 & 13.07 & 8.80 \\
Meituan-Robotics-0~\citep{robodojoleaderboard2026} & 19.28 & 12.67 & 8.22 & 3.67 & 16.77 & 7.75 & 29.61 & 18.58 & 10.06 & 8.89 & 4.54 & 4.25 & 14.95 & 9.53 \\
Xiaomi-Robotics-1~\citep{xiaomi2026xiaomirobotics1} & \best{35.65} & \best{28.00} & \best{11.44} & \best{6.00} & 26.69 & 18.83 & 38.39 & 23.67 & 7.81 & 6.56 & 3.94 & 3.58 & 20.07 & 13.93 \\
GalaxeaVLA (G0.5)~\citep{liu2026g05} & \underline{27.72} & \underline{21.44} & \underline{9.20} & \underline{4.22} & \best{28.25} & \best{20.42} & \best{44.12} & \best{32.25} & 8.61 & 7.33 & 1.73 & 1.58 & 20.23 & 14.88 \\
DM0.5~\citep{dexmal2026dm05} & 23.49 & 17.89 & 8.06 & 4.00 & 24.82 & 16.75 & 33.70 & 19.50 & \best{47.74} & \best{47.44} & \underline{2.43} & \underline{2.08} & \best{24.90} & \best{19.34} \\
\midrule
\multicolumn{15}{c}{\textbf{WAM}} \\
\midrule
OpenWAM-$\alpha$~\citep{wang2026openwam} & \underline{33.16} & \underline{25.56} & \underline{8.26} & \underline{4.11} & \underline{18.45} & \underline{9.25} & \underline{34.93} & \underline{25.33} & \underline{10.41} & \underline{9.11} & \underline{1.41} & \underline{1.08} & \underline{17.18} & \underline{11.92} \\
\textbf{\model{} (Ours)} & \textbf{39.08} & \textbf{33.33} & \textbf{22.97} & \textbf{16.33} & \textbf{30.96} & \textbf{23.00} & \textbf{43.75} & \textbf{31.50} & \textbf{18.48} & \textbf{16.67} & \textbf{22.74} & \textbf{22.00} & \textbf{29.39} & \textbf{23.60} \\
\bottomrule
\end{tabular}%
}
\end{table}

\section{Analysis}
\label{sec:analysis}
\subsection{Inference Cost}
\label{sec:cost}
Sharing the transformer between the future frame and the actions makes inference slower.
We time one call that maps an observation to a chunk of 16 actions with 20 solver steps, using the same GPU and input for both models, and average ten runs after two warm-up runs.
\model{} takes 0.446\,s per call, 1.87 times the 0.239\,s of the dual-expert control.
Both models use the same text encoder and image autoencoder, so the difference arises in the transformer.

The dual-expert control runs its visual transformer once, on the instruction and the current frame with the noise level set to zero, and stores the resulting keys and values.
Each of the 20 solver steps then updates only the 16 action tokens in its smaller action expert.
\model{} instead passes the full sequence of 721 tokens (Appendix~\ref{app:attention}) through the 4B transformer at every solver step.
The prefix cannot be reused across steps, because FLUX.2 modulates every token, including those of the prefix, with the current noise level.
The cost of \model{} therefore grows much faster with the number of solver steps than that of the control, and its smaller number of trainable parameters does not make inference faster.
In the RoboTwin evaluation, the policy is queried once every 16 executed actions with 10 solver steps, a setting we did not time.
A model trained with its prefix conditioned on a fixed noise level could cache the prefix as the control does, and step distillation \citep{akbari2026flashwam} could reduce the number of solver steps; we have tested neither.

\subsection{Mechanism Controls}
\label{sec:mechanisms}
Two controls test whether the interface must be learned and whether the noisy action and future-frame tokens need to attend to each other, and three further variants examine the shared noise level.
The two controls are trained under the full-data protocol for the same 104,850 updates as the run in Appendix~\ref{app:training-curve} and evaluated with 100 episodes per task and condition; the fixed codec reaches 94.57\% in this setting.

\paragraph{Learned interface.}
The dual-expert comparison removes a large action-specific path and changes token interaction at the same time.
To isolate the interface, the learned-linear control replaces the fixed codec with a trainable $A\!\rightarrow\!D\!\rightarrow\!A$ encoder--decoder pair, initialized to reproduce the fixed mapping, while retaining the shared transformer; for the 14-dimensional RoboTwin actions, it adds 3,726 trainable parameters, including biases.
It reaches 94.50\% on average, 0.07 percentage points below the fixed codec; it succeeds more often in clean scenes and less often in randomized scenes (Table~\ref{tab:learned-linear}).
A learned interface therefore brings no measurable gain in this setting, and we keep the fixed codec because it adds no trainable parameters and needs no initialization.
\begin{table}[htbp]
\centering
\caption{Fixed versus learned-linear action interfaces on RoboTwin~2.0 under the full-data protocol (SR, percent), with 100 evaluation episodes per task and condition.}
\label{tab:learned-linear}
\small
\begin{tabular}{lrrr}
\toprule
Interface & Clean & Randomized & Average\\
\midrule
Fixed (\model{}) & 94.68 & \best{94.46} & \best{94.57}\\
Learned-linear & \best{95.02} & 93.98 & 94.50\\
\midrule
$\Delta$ & $+0.34$ & $-0.48$ & $-0.07$\\
\bottomrule
\end{tabular}
\end{table}

\paragraph{Noisy action--image visibility.}
The isolated-attention control keeps the sequence, the objective, and the codec of \model{} but prevents the noisy action tokens and the noisy future-frame tokens from attending to one another; each group still attends to itself and to the prefix (Appendix~\ref{app:attention}).
Removing this visibility reduces the average success rate by 0.58 percentage points, from 94.57\% to 93.99\%, with drops in both clean scenes (0.76 points) and randomized scenes (0.40 points; Table~\ref{tab:iso-attn}).
The interaction between the two token groups may therefore contribute a small gain, which a single training run per model cannot establish.
\begin{table}[htbp]
\centering
\caption{\model{} against the isolated-attention control on RoboTwin~2.0 under the full-data protocol (SR, percent), with 100 evaluation episodes per task and condition.}
\label{tab:iso-attn}
\small
\begin{tabular}{lrrr}
\toprule
Method & Clean & Randomized & Average\\
\midrule
\model{} & \best{94.68} & \best{94.46} & \best{94.57}\\
Isolated attention & 93.92 & 94.06 & 93.99\\
\midrule
$\Delta$ & $-0.76$ & $-0.40$ & $-0.58$\\
\bottomrule
\end{tabular}
\end{table}

\paragraph{Decoupled noise levels.}
\model{} applies one noise level to the future-frame and action tokens of a sample (Section~\ref{sec:objective}).
Three variants built on Self-Flow \citep{chefer2026selfflow} depart from this choice and are trained and evaluated in the setting of Table~\ref{tab:sparse}.
The first adopts the dual-timestep scheduling of Self-Flow, which assigns different noise levels to different tokens, together with a representation loss from an exponential-moving-average (EMA) teacher.
The second additionally structures the timestep masks by modality, and the third also withholds the action labels of 25\% of the samples and replaces them with labels generated by the teacher.
Their average success rates are 88.7\%, 88.3\%, and 85.5\%, compared with 88.0\% for the shared noise level (Appendix~\ref{app:self-flow}).
The first two variants succeed more often in clean scenes and less often in randomized scenes, and their averages exceed that of \model{} by at most 0.7 percentage points, a difference comparable to the uncertainty of a 10-episode evaluation.
The third variant reduces the average by 2.5 points.
Because every variant also changes the training objective, these results do not isolate the effect of the noise levels, but they give no indication that separate noise levels improve on the shared one in this setting.

\section{Limitations and Discussion}
\label{sec:limitations}
\paragraph{Scope of the architectural evidence.}
The matched comparison shows that a separate action expert is not required in this sampling regime, but it does not establish that every dedicated action head is unnecessary.
Relative to the dual-expert control, \model{} changes action-specific capacity, token visibility, and the processing path together.
The controls of Section~\ref{sec:mechanisms} separate two of these factors under the full-data protocol: a learned linear interface performs on par with the fixed codec, and removing the visibility between action and future-frame tokens costs 0.58 percentage points.
Because they were run in a different data regime, they do not decompose the 9.6-point gap of the matched comparison, and no comparison in this paper includes independent training seeds.
Joint denoising allows interaction between visual and action predictions, and the isolated-attention control suggests a small gain from this interaction, but neither establishes that the predicted future frame informs action selection.
At the highest noise level, the future-frame tokens contain no information about the observed future, and a single future frame can be consistent with several action sequences.
Additional conditioning information can reduce the squared error of an ideal predictor, although this does not imply higher closed-loop success for a trained model.
Cross-modal coupling and the reuse of pretrained weights are therefore motivations for the design rather than established explanations of its performance.

\paragraph{Inference and deployment.}
Because future-frame and action tokens are updated at every solver step, one inference call with 20 solver steps takes 1.87 times as long as in the dual-expert control.
The 10-step setting used for evaluation was not timed.
Part of this cost follows from the architecture, which passes the future-frame tokens through the full transformer at every solver step.
In addition, unlike the dual-expert control, \model{} recomputes the conditioning prefix at every solver step, because FLUX.2 modulates the prefix with the current noise level (Section~\ref{sec:cost}).
All experiments are conducted in simulation, and the results do not establish real-time operation, performance on physical robots, or safe deployment.

\section{Conclusion}
We investigated whether continuous control requires a separate action expert, or whether the visual generative pathway of a pretrained image model can be adapted to predict actions directly.
\model{} maps each action step to a token in the visual latent space through a fixed codec, and a shared transformer denoises these tokens together with the tokens of a future frame, without a trainable action encoder, an action-specific output projection, or a separate action expert.
Under matched data and optimization settings on RoboTwin~2.0, with one of every 20 training-window starts retained, \model{} achieves 88.0\% average success compared with 78.4\% for the dual-expert control and reaches the final success rate of the control with fewer training updates.
Across benchmarks with different data regimes, it reaches 96.12\% average success on RoboTwin~2.0 with augmented demonstrations and 91.8\% on LIBERO-Plus with augmented training data, and its RoboDojo Score of 29.39 is the highest among the leaderboard entries of September 15, 2026.
These results show that the fixed interface works in the evaluated settings, although joint denoising makes inference slower than in the dual-expert control (Section~\ref{sec:cost}).
Two controls under the full-data protocol show that a learned linear interface does not improve on the fixed codec and that removing the visibility between action and future-frame tokens costs 0.58 percentage points.
Identifying the factors behind the advantage over the dual-expert control still requires further controls in the matched setting and independent training seeds, and all results remain to be tested on physical robots.

\section*{Acknowledgments}
We thank Awomo for providing the computing resources used for training and evaluation.

\section*{Reproducibility Statement}
Section~\ref{sec:method} defines the Action-as-Patch codec, the training objective, the token sequence and attention mask, and the inference procedure, and Appendix~\ref{app:attention} shows the sequence and the mask for one RoboTwin~2.0 sample.
Section~\ref{sec:setup} describes the benchmarks, the training data of each protocol, the matched control, and the evaluation protocol.
Appendix~\ref{app:implementation} lists the settings of the training runs: the model settings that the runs share, the global batch size and number of epochs of each run, the filtering of the training data, the image augmentation of each protocol, and the inputs and text encoder of the RoboDojo run.
It also describes the architecture, initialization, and parameter count of the dual-expert control, and it gives the evaluation seeds of RoboTwin~2.0 and LIBERO, the number of LIBERO trials, and the replanning interval of each benchmark.
Appendix~\ref{app:training-curve} reports the success rates of intermediate checkpoints of one complete training run.
No reported result is repeated with independent training or evaluation seeds.

\section*{AI Use Statement}
AI assistants were used to inspect implementation files, organize experimental results, and draft and edit the text of this paper.
The authors reviewed the numerical results, citations, and claims, and they take full responsibility for the content.

\bibliography{references}
\bibliographystyle{iclr2027_conference}
\clearpage
\appendix
\section{LIBERO-Pro Comparison}
\label{app:additional}
LIBERO-Pro \citep{zhou2025liberopro} evaluates policies trained on the original LIBERO data under changes to the objects, to their initial positions, to the wording of the instruction, and to the task itself.
Table~\ref{tab:libero-pro-comparison} gives the success rate under each perturbation and on the unperturbed tasks, averaged over the four LIBERO suites.
The results of $\pi_0$, OpenVLA, and $\pi_{0.5}$ are taken from \citet{zhou2025liberopro}, and those of SimVLA and QuoVLA from their papers.

\model{} reaches an average success rate of 57.9\%, above $\pi_0$ and OpenVLA and below the other three listed methods.
On the unperturbed tasks and under paraphrased instructions, it succeeds in 98.4\% and 96.8\% of the trials, 1.4 and 2.7 percentage points below the best listed results.
Under object perturbations, its success rate of 80.8\% is the lowest among the listed methods.
Changes to the initial positions and to the task are difficult for every listed method: \model{} reaches 8.8\% and 5.0\%, and only QuoVLA exceeds 21\% in either category.
\begin{table}[H]
\centering
\caption{LIBERO-Pro results by perturbation (SR, percent), averaged over the four LIBERO suites. Original denotes the unperturbed tasks. Methods are sorted by average SR, and \model{} is shown in bold; averages are computed before rounding.}
\label{tab:libero-pro-comparison}
\footnotesize
\setlength{\tabcolsep}{4pt}
\begin{tabular}{lcccccc}
\toprule
Method & Original & Object & Position & Semantic & Task & \textbf{Average} \\
\midrule
$\pi_0$~\citep{black2024pi0} & 92.3 & 90.5 & 0.0 & 90.5 & 0.0 & 54.7 \\
OpenVLA~\citep{kim2024openvla} & 97.0 & 93.0 & 0.0 & 97.3 & 0.0 & 57.5 \\
SimVLA~\citep{luo2026simvla} & 98.5 & 81.5 & 8.3 & 98.8 & 6.0 & 58.6 \\
$\pi_{0.5}$~\citep{intelligence2025pi05} & 96.5 & 96.0 & 20.8 & 95.8 & 0.8 & 62.0 \\
QuoVLA~\citep{wang2026quovla} & 99.8 & 88.3 & 36.0 & 99.5 & 25.5 & 69.8 \\
\midrule
\textbf{\model{} (Ours)} & \textbf{98.4} & \textbf{80.8} & \textbf{8.8} & \textbf{96.8} & \textbf{5.0} & \textbf{57.9} \\
\bottomrule
\end{tabular}
\end{table}

\FloatBarrier

\section{Training Dynamics under the Full-Data Protocol}
\label{app:training-curve}
\begin{figure}[htbp]
  \centering
  \includegraphics[width=\textwidth]{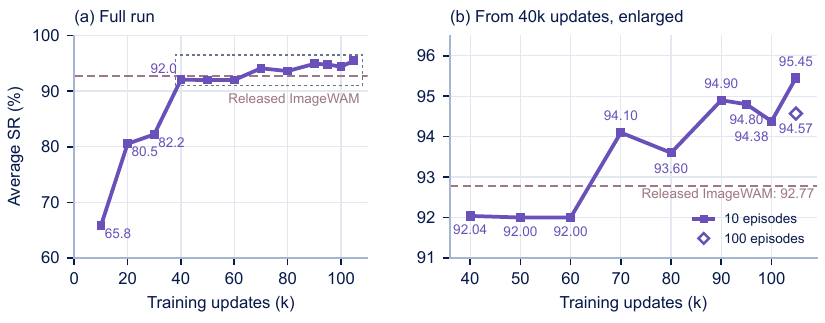}
  \caption{Success rate of \model{} over a training run under the full-data protocol without augmented demonstrations. \textbf{(a)}~The full run of 104,850 updates. \textbf{(b)}~The checkpoints from 40k updates onward, enlarged. Squares denote evaluations with 10 episodes per task and condition, the open diamond the 100-episode evaluation of the final checkpoint, and the dashed line the 100-episode evaluation of the released ImageWAM checkpoint.}
  \label{fig:full-curve}
\end{figure}

Figure~\ref{fig:full-curve} traces the success rate of \model{} over one training run under the full-data protocol.
The run uses the clean and randomized demonstrations without the augmented demonstrations and trains for five epochs, or 104,850 updates with 256 samples per update.
The checkpoints at every 10k updates, at 95k updates, and at the end of training are evaluated with 10 episodes per task and condition, and the final checkpoint is also evaluated with 100 episodes.
The dashed line marks our 100-episode evaluation of the released ImageWAM checkpoint, which reaches 92.77\%, slightly below the 93.38\% reported for it in Table~\ref{tab:robotwin-full}.

Most of the improvement occurs within the first two epochs: the success rate rises from 65.8\% after 10k updates to 80.5\% after 20k updates and 92.0\% after 40k updates.
It then stays at 92.0\% until 60k updates.
From 70k updates onward, every evaluated checkpoint exceeds the released ImageWAM checkpoint, although the intermediate checkpoints are evaluated with fewer episodes.
Between 70k updates and the end of training, the success rate varies between 93.60\% and 95.45\%.
Differences of this size between neighboring checkpoints are comparable to the uncertainty of a 10-episode evaluation: the final checkpoint reaches 95.45\% with 10 episodes and 94.57\% with 100 episodes per task and condition.
For the 10-episode value, two of the 100 task--condition units use their 100-episode results because their 10-episode evaluations did not complete.
Under the 100-episode evaluation, the final checkpoint exceeds the released ImageWAM checkpoint by 1.80 percentage points.
The 96.12\% in Table~\ref{tab:robotwin-full} comes from a separate configuration, which adds the 2,100 augmented demonstrations and trains for 110k updates; it exceeds the 94.57\% of the run traced in Figure~\ref{fig:full-curve} by 1.55 points.
\FloatBarrier

\section{Per-Task Results under the Clean-to-Random Protocol}
\label{app:c2r-per-task}
\begin{figure}[htbp]
  \centering
  \includegraphics[width=\textwidth]{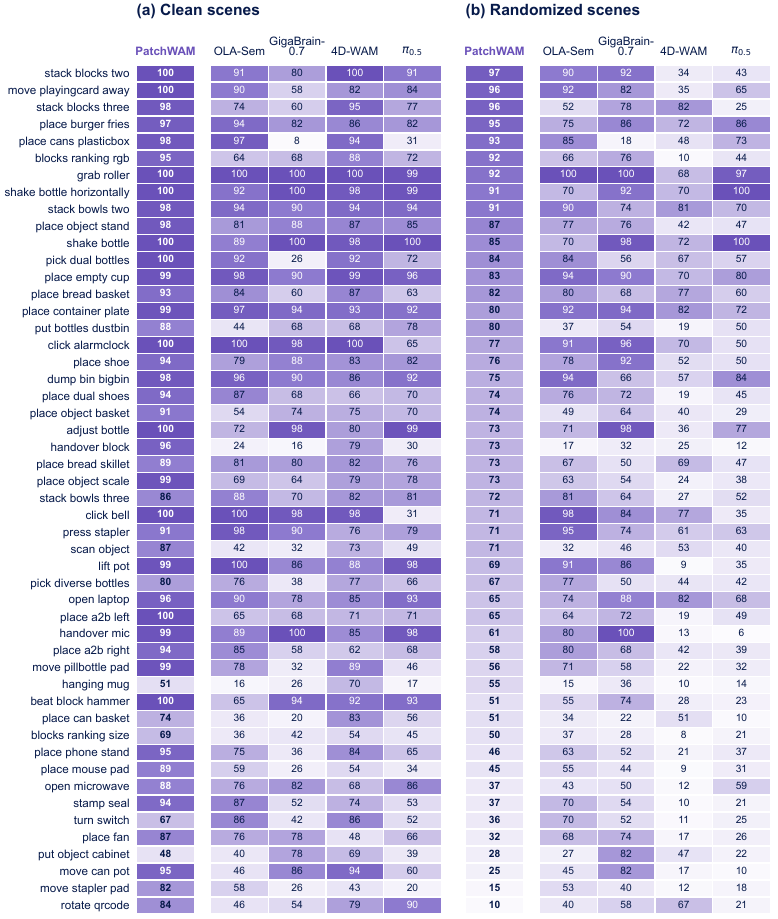}
  \caption{Per-task success rates (percent) under the clean-to-random protocol for \model{} at the reported checkpoint and the four strongest baselines of Table~\ref{tab:robotwin-c2r}, all evaluated with 100 episodes per task and condition. Tasks are sorted by the success rate of \model{} in randomized scenes. \textbf{(a)}~Clean scenes. \textbf{(b)}~Randomized scenes.}
  \label{fig:c2r-per-task}
\end{figure}

Figure~\ref{fig:c2r-per-task} breaks down the clean-to-random results of Table~\ref{tab:robotwin-c2r} by task.
It compares \model{} at the reported checkpoint with the four strongest baselines of that table, whose per-task results are taken from the leaderboard \citep{robotwinleaderboard2026}; every method is evaluated with 100 episodes per task and condition.
In clean scenes, \model{} succeeds in every episode of 11 tasks and falls below 80\% on five tasks, the two lowest being putting the object into the cabinet at 48\% and hanging the mug at 51\%.
In randomized scenes, it reaches at least 80\% on 16 tasks and at most 20\% on two, rotating the QR code and moving the stapler to the pad.
Compared with OLA-Sem and GigaBrain-0.7, whose success rates in randomized scenes are close to its own, \model{} succeeds more often in randomized scenes on 23 and 24 tasks, respectively, and less often on 26 tasks each, so policies with similar success rates differ in which tasks they solve.
\FloatBarrier

\section{Qualitative Clean-to-Random Rollouts}
\label{app:c2r-qualitative}
\begin{figure}[!htbp]
  \centering
  \includegraphics[width=\textwidth]{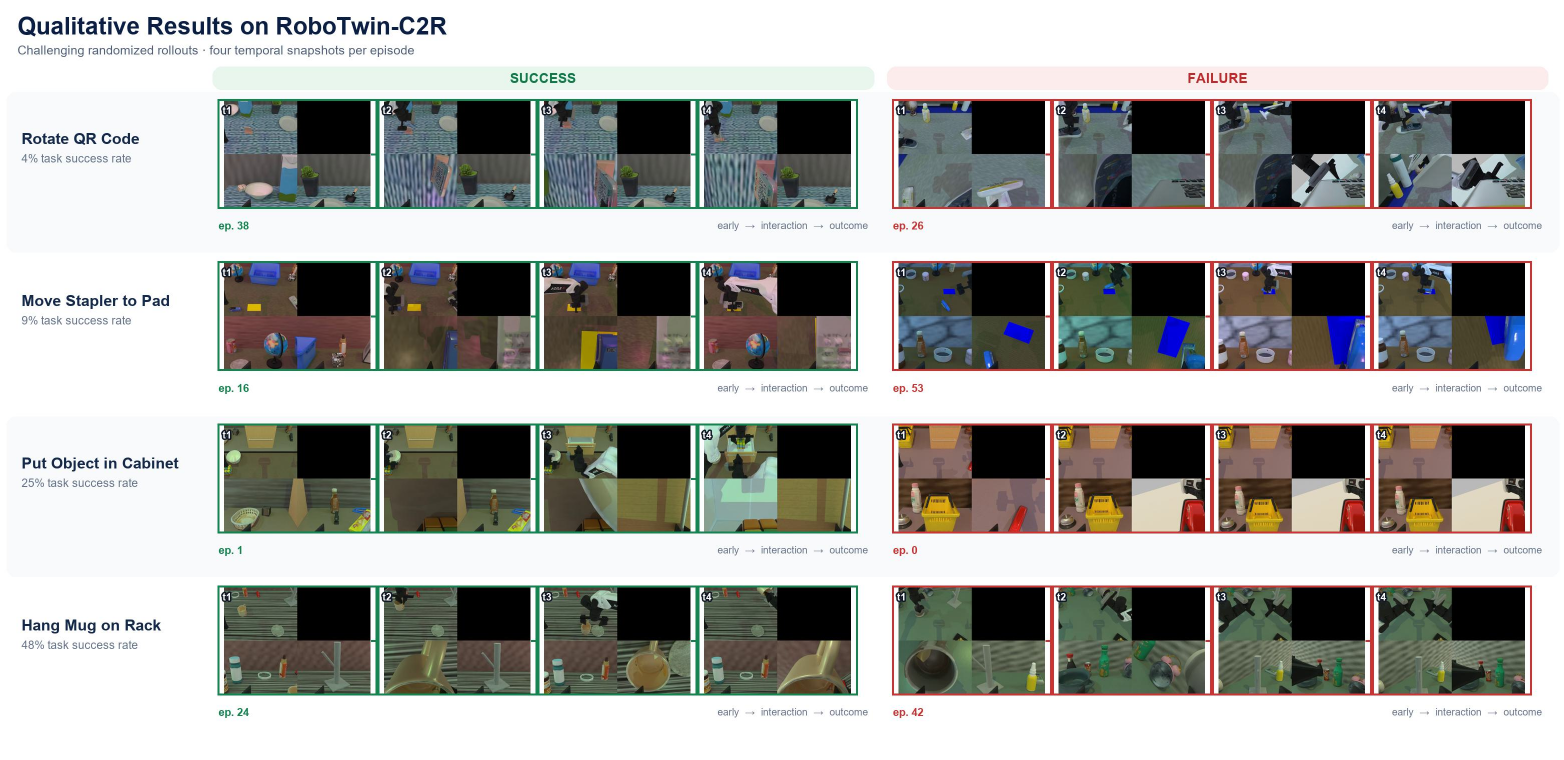}
  \caption{\textbf{Qualitative clean-to-random rollouts in randomized scenes.}
  Each row pairs a successful rollout (green) with a failed rollout (red) from the same task, with four frames ordered from early interaction to the final outcome.
  The four tasks have success rates of 4\%, 9\%, 25\%, and 48\% in the evaluation of this checkpoint, so each successful rollout comes from a task that the model completes in at most half of the episodes.}
  \label{fig:c2r-qualitative}
\end{figure}

Figure~\ref{fig:c2r-qualitative} shows paired outcomes from four difficult tasks in randomized scenes.
The examples are selected from all 100 evaluation episodes per task of the clean-to-random \model{} checkpoint after 100k updates.
Success and failure labels are produced by the RoboTwin evaluator rather than assigned manually.

\FloatBarrier

\section{Token Sequence and Attention Mask}
\label{app:attention}
\begin{figure}[t]
  \centering
  \includegraphics[width=\textwidth]{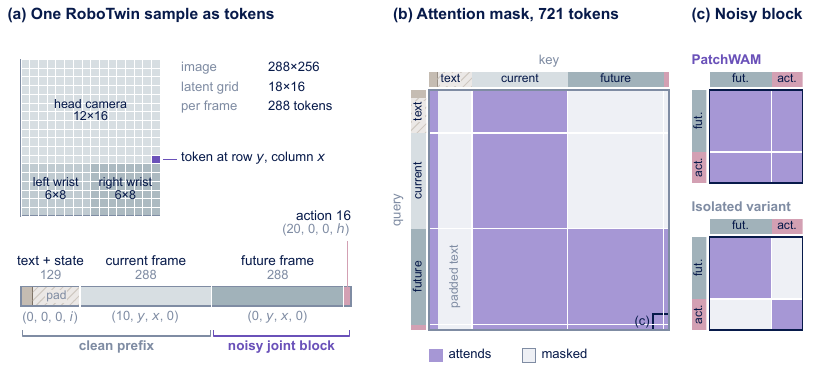}
  \caption{\textbf{Token sequence and attention mask for one RoboTwin sample.}
  \textbf{(a)}~Each frame becomes an $18\times16$ grid of latent tokens; the bar shows the full sequence to scale, with the positional coordinates of each group.
  \textbf{(b)}~The attention mask over all 721 tokens, with rows as queries and columns as keys; the strips along the edges mark the token groups of~(a).
  \textbf{(c)}~The last 48 tokens of the noisy block, boxed in~(b), enlarged for \model{} and for the isolated-attention control of Section~\ref{sec:mechanisms}.
  The sample is taken from the first training episode, with the instruction ``Grab the smooth green plastic bottle and lift it with the left arm'', whose prompt occupies 25 of the 128 text tokens.}
  \label{fig:attention-case}
\end{figure}
Figure~\ref{fig:attention-case} illustrates Section~\ref{sec:attention} with one RoboTwin sample.
The three camera views are composed into a $288\times256$ image, with the head camera above the left and right wrist cameras, and the image autoencoder maps this image to an $18\times16$ grid of latent tokens.
Each frame therefore contributes 288 tokens.
The instruction is padded to 128 text tokens, and the proprioceptive token is inserted directly after the valid text, so that the text block contains 129 tokens.
With the two frames and the 16 action tokens, the sequence contains 721 tokens, of which the action tokens account for about 2\%.

The positional coordinates follow the convention of Section~\ref{sec:attention}.
Text tokens carry their sequence index on the fourth axis, image tokens carry their row and column on the second and third axes, and action tokens carry their step index on the fourth axis.
The first axis separates the current frame (10), the future frame (0), and the actions (20).

Panel~(b) shows the mask used in training and inference.
Prefix queries attend only to prefix keys, whereas future-frame and action queries attend to every non-padded key.
Padded text tokens are masked as keys for all queries.
Panel~(c) contrasts this mask with the isolated-attention control, in which future-frame and action tokens no longer attend to each other but retain access to the prefix and to their own group.

\FloatBarrier
\section{Implementation Details}
\label{app:implementation}
This appendix gives the settings of the training runs and evaluations in this paper.
Table~\ref{tab:shared-settings} lists the settings that the runs share, and Table~\ref{tab:run-settings} gives the number of GPUs, the batch sizes, and the number of epochs of each run.
The RoboDojo run also uses past frames and a vision-language (VL) model as its text encoder, and the paragraphs \emph{RoboDojo run} and \emph{RoboDojo text encoder} below describe its configuration.

\begin{table}[htbp]
\centering
\caption{Settings shared by the training runs. The RoboDojo run uses a different text encoder and an additional loss, as described below.}
\label{tab:shared-settings}
\small
\begin{tabular}{ll}
\toprule
Setting & Value\\
\midrule
Transformer & FLUX.2 [klein] 4B Base, all weights trained\\
State projection & linear layer, trained\\
Image autoencoder & FLUX.2 autoencoder, frozen\\
Text encoder & Qwen3-4B, frozen, 128 instruction tokens\\
Action chunk & $H=16$ steps, future frame at the end of the chunk\\
Action-as-Patch & token width $D=128$, scale $\alpha=1$\\
Noise-level shift & $s=5$ in training and inference\\
Loss weights & $\lambda_x=0.5$, $\lambda_u=1.0$\\
\bottomrule
\end{tabular}
\end{table}

\begin{table}[htbp]
\centering
\caption{Parallelism, batch size, and number of epochs of each training run. The global batch size is the product of the number of GPUs, the gradient-accumulation steps, and the batch size per GPU. The clean-to-random and RoboDojo runs stop after a fixed number of updates instead of a number of epochs.}
\label{tab:run-settings}
\small
\begin{tabular}{llrrrrr}
\toprule
& & & & \multicolumn{2}{c}{Batch size} & \\
\cmidrule(lr){5-6}
Benchmark & Run & GPUs & Accumulation & Per GPU & Global & Epochs\\
\midrule
RoboTwin~2.0 & Full-data & 32 & 2 & 4 & 256 & 5\\
& Full-data, augmented & 32 & 2 & 4 & 256 & 5\\
& Matched comparison & 8 & 2 & 4 & 64 & 10\\
& Clean-to-random & 16 & 1 & 16 & 256 & ---\\
LIBERO & Original data & 16 & 1 & 8 & 128 & 10\\
LIBERO-Plus & Augmented data & 16 & 2 & 8 & 256 & 10\\
RoboDojo & Past frames, VL text encoder & 32 & 1 & 8 & 256 & ---\\
\bottomrule
\end{tabular}
\end{table}

\paragraph{Training data.}
Each training sample is a window of 17 time steps that covers the current observation, the 16 actions of the chunk, and the observation at the end of the chunk; apart from the past frames of the RoboDojo run, only the first and the last image of the window are loaded.
On RoboTwin~2.0, idle segments, in which the commanded action matches the current robot state, are removed from each demonstration before the windows are formed.
Idle steps at the end of a demonstration are kept, because they often contain the release of the object.
After this step, the RoboTwin~2.0 full-data set contains about 5.37 million training windows.
The matched comparison keeps one window start out of every twenty, or about 268,000 windows, and both of its runs pass over each retained window ten times.
The RoboDojo run keeps all steps of each demonstration.
On LIBERO, the four suites are used without their no-op actions.

\paragraph{Image augmentation.}
RoboTwin~2.0 and RoboDojo runs other than the clean-to-random run augment half of the training samples online.
An augmented sample is cropped to between 95\% and 100\% of each image side and rotated by up to 5 degrees, and it additionally receives color changes, Gaussian noise with a standard deviation of 0.01, or both.
The color changes scale brightness, contrast, and saturation by factors between 0.8 and 1.2, shift the hue by up to 11 degrees, and apply a gamma between 0.9 and 1.1.
All camera views and both frames of a sample share the same crop, rotation, and color parameters.
In the RoboDojo run, the parameters are drawn separately for each camera view and for the past frames, and the input images of the text encoder are not augmented.
LIBERO runs apply the same transforms with wider ranges, plus an exposure change, to 65\% of the samples: crops to at least 92\% of each side, rotations of up to 8 degrees, brightness and contrast factors between 0.7 and 1.3, saturation factors between 0.75 and 1.25, hue shifts of up to 14 degrees, gammas between 0.8 and 1.25, exposure changes of up to 0.25 stops, and noise with a standard deviation of 0.015.
The clean-to-random run replaces these transforms with a stronger randomization, which is applied to 80\% of the samples, with parameters drawn separately for each camera view and shared by both frames.
It scales brightness, contrast, and saturation by factors between 0.6 and 1.4, shifts the hue by up to 36 degrees, applies a gamma between 0.7 and 1.4 and an exposure change of up to 0.5 stops, scales the red and blue channels by up to 15\% to change the color temperature, blurs the image with probability 0.3, and adds Gaussian noise with a standard deviation of 0.03.
With probability 0.5 each, it then applies an AdaIN transform, which moves the mean and standard deviation of each color channel toward random targets, and an FFT-based transform, which scales the amplitude spectrum of the image by random factors between 0.5 and 1.5 while keeping its phase.
Neither of these two transforms moves edges or objects, and the run uses no crops or rotations.

\paragraph{RoboDojo run.}
The RoboDojo run keeps the three camera views as separate $256\times256$ images, so each view contributes 256 latent tokens per frame, and, as on RoboTwin~2.0, both actions and proprioceptive states are 14-dimensional.
The future frames of all three views are predicted, and the views are distinguished by their values on the first positional axis, which are 10, 11, and 12 for the current frame and 0, 1, and 2 for the future frame.
The run also conditions on 20 slots of past head-camera frames, one per second over the last 20 seconds.
At inference, slot $k$ holds the frame from $k$ seconds before the current step, and slots that lie entirely before the start of the episode are masked.
During training, the frame of each slot is drawn uniformly within 0.4 seconds of this time, the whole history is dropped with probability 0.2 and each slot with probability 0.2, and a random number of the most recent remaining slots, between 0 and 20, is kept.
Each past frame is encoded by the image autoencoder, average-pooled to $4\times4$ tokens, and added to the prefix with the value $-k$ on the first positional axis, so the history adds 320 tokens and no parameters.

\paragraph{RoboDojo text encoder.}
The text encoder of the RoboDojo run is Qwen3-VL-4B-Instruct \citep{bai2025qwen3vl}.
It reads the instruction, the current head-camera image, and up to 20 past head-camera frames, one per second over the last 20 seconds, all resized to $448\times448$ and not augmented.
During training, each past frame is shifted in time by up to one video frame; unlike the past frames of the transformer, none is dropped.
The vision encoder maps each image to 196 tokens, and the tokens of each past frame are average-pooled to $4\times4$ before they enter the language model.
The transformer receives the concatenated hidden states of layers 9, 18, and 27 at the instruction tokens and those of layers 18, 27, and 36 at the last prompt token.
The vision encoder of the VL model stays frozen.
LoRA adapters \citep{hu2022lora} with rank 64, a scaling factor of 2, and dropout 0.05 are trained on the attention and feed-forward projections of its language model; the adapters add 132.1M trainable parameters.
Subtask labels divide each demonstration into segments and describe each segment in one sentence, and every training demonstration carries such labels.
The subtask of the current step is appended after the prompt and predicted with a cross-entropy loss of weight 0.1 and label smoothing 0.1.
Because the language model attends causally, the appended subtask does not change the hidden states that the transformer receives, and the subtask is not used at inference.

\paragraph{Dual-expert control.}
The dual-expert control is ImageWAM \citep{zhang2026imagewam} with the same FLUX.2 [klein] 4B backbone.
Its action expert has 5 double-stream and 20 single-stream blocks of width 1024; the block counts and the attention dimensions (24 heads of size 128) match those of the backbone, so that the action tokens join the attention of the backbone in every block.
The action expert embeds each action step with a linear layer, reads out the action velocity with a linear layer, and starts from random weights.
It adds 0.64B trainable parameters to the model.

\paragraph{Evaluation.}
RoboTwin~2.0 evaluations use seed 43 and the unseen instructions of the benchmark, and the policy executes each predicted chunk of 16 actions in full before it predicts the next chunk from the new observation.
On RoboDojo, the policy replans after executing 8 of the 16 actions.
LIBERO evaluations run 20 trials per task with seed 42.
On LIBERO and LIBERO-Plus, the policy predicts 16 actions and replans after executing 12 of them.
All evaluations use 10 solver steps and no classifier-free guidance.

\section{Supplementary Self-Flow Results}
\label{app:self-flow}
\begin{table}[H]
\centering
\caption{Self-Flow variants after 41,940 updates (SR, percent), with one of every 20 window starts retained and 10 episodes per task and condition.}
\small
\begin{tabular}{lrrr}
\toprule
Configuration & Clean & Randomized & Average\\
\midrule
Dual-expert control & 77.2 & 79.6 & 78.4\\
\model{} & 87.6 & \best{88.4} & 88.0\\
Self-Flow variant 1 & \best{90.0} & 87.4 & \best{88.7}\\
Self-Flow variant 2 & 89.4 & 87.2 & 88.3\\
Self-Flow variant 3 & 85.4 & 85.6 & 85.5\\
\bottomrule
\end{tabular}
\end{table}
Variant 1 combines dual-timestep scheduling and an EMA-teacher representation loss.
Variant 2 adds modality-structured timestep masks.
Variant 3 also withholds the action labels of 25\% of the samples and uses labels generated by the teacher instead.
These variants change the training objective and are therefore not independent training repetitions of \model{}.
Variants 1 and 2 succeed more often in clean scenes, and all three succeed less often in randomized scenes.

\end{document}